%% file: bmvc_final.tex
\documentclass{bmvc2k}

\pdfoutput=1

\title{SurReal: enhancing Surgical simulation Realism using style transfer}

\addauthor{Imanol Luengo}{imanol.luengo@touchsurgery.com}{1}
\addauthor{Evangello Flouty}{evangello.flouty@touchsurgery.com}{1}
\addauthor{Petros Giataganas}{petros.giataganas@touchsurgery.com}{1}
\addauthor{Piyamate Wisanuvej\textsuperscript{1,2,}}{piyamate@touchsurgery.com}{3}
\addauthor{Jean Nehme}{jean@touchsurgery.com}{1}
\addauthor{Danail Stoyanov\textsuperscript{1,}}{danail.stoyanov@touchsurgery.com}{2}

\addinstitution{
 Digital Surgery\\
 London, UK
}

\addinstitution{
Wellcome / EPSRC Centre for Interventional and Surgical Sciences\\
University College London\\
London, UK
}

\addinstitution{
 Kasetsart University\\
 Bangkok, Thailand
}

\usepackage{bm}
\usepackage{tikz}

\runninghead{SurReal}{Enhancing Surgical simulation Realism}

\newcommand{\norm}[1]{\left\lVert#1\right\rVert}

\begin{document}

\maketitle

\begin{abstract}

Surgical simulation is an increasingly important element of surgical education. Using simulation can be a means to address some of the significant challenges in developing surgical skills with limited time and resources. The photo-realistic fidelity of simulations is a key feature that can improve the experience and transfer ratio of trainees. In this paper, we demonstrate how we can enhance the visual fidelity of existing surgical simulation by performing style transfer of multi-class labels from real surgical video onto synthetic content. We demonstrate our approach on simulations of cataract surgery using real data labels from an existing public dataset. Our results highlight the feasibility of the approach and also the powerful possibility to extend this technique to incorporate additional temporal constraints and to different applications.

\end{abstract}

\section{Introduction}
\label{sec:intro}

Surgical skills are traditionally learned by trainees using the apprenticeship model, through observation, mentoring and gradually practicing on patients \cite{reznick2006teaching}. As the complexity of operations, devices and operating rooms has increased with modern imaging and robotics technology, more effective and efficient training systems are necessary. Surgical simulation offers a potential solution to training needs and can be used to good effect in a low-stress environment without risking the patients' safety. To be effective, simulation should be realistic both in visual fidelity and in functional and behavioral features of anatomical structures. 

In addition to offering new methods for surgical training, digital simulation tools can also be used to offer new capabilities like procedural rehearsal that can be used for \textit{in situ} practice \cite{Kowalewski2017}. Combined with patient or procedure specific information of anatomical models from tomographic scans, platforms for rehearsal could be used to ease the challenges of very difficult cases or to ensure optimal performance. To enable this capability, realism is critical and merging information from pre-built simulation environments together with information, such as video, from the site during surgery, can be an approach to achieving high realism \cite{haouchine2017dejavu}. 

While significantly improved with modern graphics techniques, the photo-realistic fidelity of virtual simulations in surgery is still limited. This is a hurdle to the overall life-like experience for the trainees . Tissue and surgical lighting modelling, and accurate representation of various layers for different anatomical structures can be particularly challenging and computationally expensive to generate. In this paper, we adopt a different, novel approach towards enhancing the visual fidelity of surgical simulations by performing label-to-label style transfer from real surgical video onto synthetic content. We demonstrate the feasibility of this method on simulated content of cataract surgery using real semantic segmentation labels from an existing public dataset \cite{bouget2017vision} (https://cataracts.grand-challenge.org/).

\subsection{Contributions}

To our knowledge this is the first time that style transfer has been used within the surgical simulation application domain. In recent years, surgical simulation has focused primarily on improving the realism of deformable tissue-instrument interactions through biomechanical modelling using finite-element techniques \cite{Sofa2007}. Our approach can be used in conjunction with deformable models to improve the photorealistic properties of simulation and can also be used to refine the visual appearance of existing systems. 

Beyond the application domain interest and novelty of the presented work, our paper reports two algorithmic contributions: (1) we generalize Whitening and Coloring Transform (WCT) by adding style decomposition, allowing the creation of "style models" from multiple style images; and (2) we introduce label-to-label style transfer, allowing region-based style transfer from style to content images, which our algorithm handles inherently with robustness to missing labels.

Additionally, we pave the way towards unlimited training data for Deep Convolutional Neural Networks (CNN) by exploiting the ability to automatically generate segmentation masks from surgical simulations. Already proven \cite{zisimopoulos2017can}, our approach will further boost the transferability by making the images more realistic.

\vspace{-2mm}
\section{Related work}

Over the last years a recent trend has appeared trying to make 3D simulations more realistic, from the 3D computer graphics point of view. The trend is being driven by the appearance of studies suggesting that some of the core skills the surgeons should have, should be learned prior to entering the OR \cite{reznick2006teaching}, which resulted in growing VR simulations and works trying to make it more realistic. Approaches range from making wet surfaces more realistic \cite{kerwin2009enhancing} to creating intra-operative enhanced simulations \cite{haouchine2017dejavu}, while at the same time recent studies \cite{smink2017realism,Kowalewski2017} validated this kind of simulation as useful pre-operative educational tools.

We propose a novel approach, different from 3D graphics, to improve the realism of a rendered simulation video by transferring the style from real surgery footage. This is driven by the recent works of Gatys et al. \cite{gatys2015neural,gatys2016image}, that push the artistic and textural transfer from one image to another to a whole different new level. Neural style-transfer can be seen as a combination of feature reconstruction and texture synthesis, as the goal is to reconstruct a whole new image with the content of an image A and the style of an alternative image B. This is achieved by designing an optimization algorithm to iteratively improve a reconstruction, minimizing the Gram Matrix (\textit{i.e.} correlation of deep features) of the style images and the feature reconstruction error of the content images. This initial approach requires, however, to solve the optimization iteratively, which takes long to render a single image. Since then, different approaches have been proposed to make it faster \cite{li2016precomputed,ulyanov2016texture} or look better \cite{huang2017arbitrary}, including recent work in photo-realistic style transfers \cite{luan2017deep}.

\begin{figure}
\vspace{-22mm}
\includegraphics[width=\textwidth,clip]{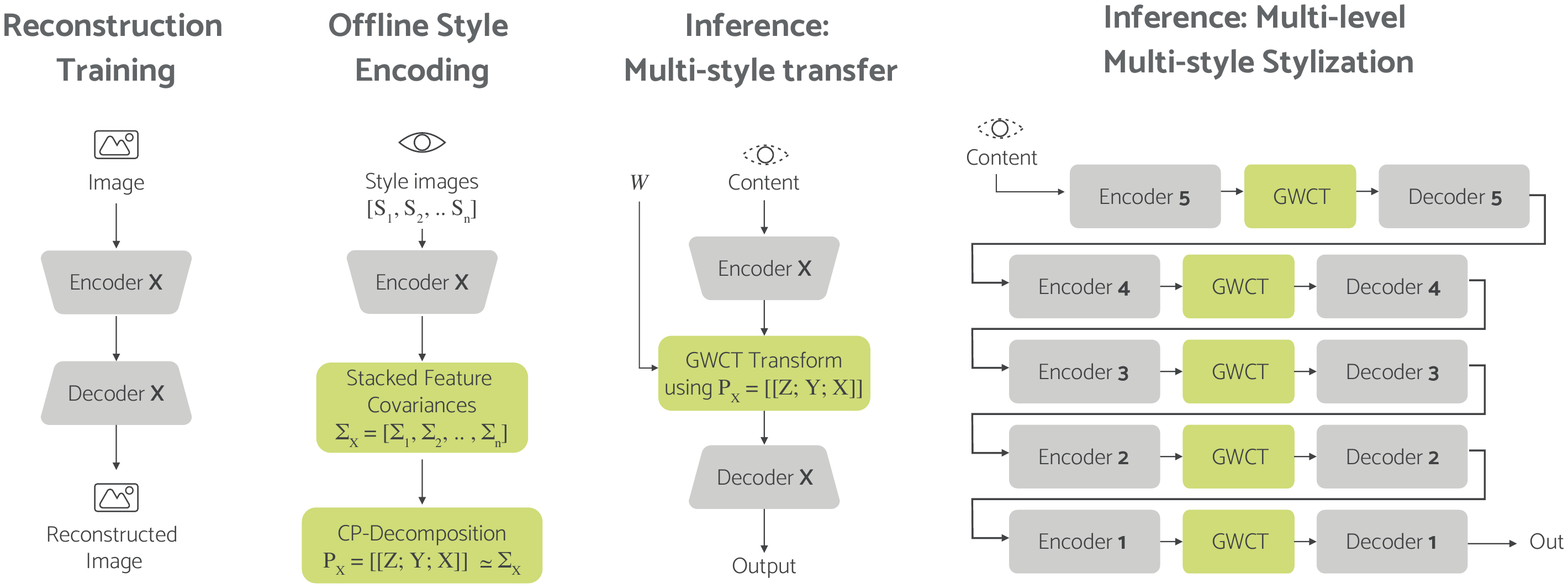}
\vspace{-26mm}
\caption{Proposed generalized multi-style transfer pipeline. From left to right: (1) As in WCT, the encoder-decoders are trained for image reconstruction. (2) The $N$ target styles are encoded offline, and a joint representation is computed using CP-decomposition. (3) In inference, the precomputed styles $P_x$ are blent using a weight vector $W$. (4) Multi-scale generalization of inference. Note that every GWCT module in (4) requires a $W$ vector, omitted for clarity.}
\label{fig:overview}
\vspace{-4mm}
\end{figure}

Our approach is more closely related to Universal Style Transfer (UST) \cite{li2017universal}, which proposes a feed-forward neural network to stylize images. Different to other feed-forward approaches \cite{chen2017stylebank,dumoulin2016learned}, UST does not require to learn a new CNN model or filters for every set of styles in order to transfer the style to a target image; instead, a stacked encoder/decoder architecture is trained solely for image reconstruction. Then, during inference of a content-style pair, a WCT is applied after both images are encoded to transfer the style from one to the other, and reconstruct only the modified image from the decoder.

We extend the work from UST by generalizing WCT. We add an intermediate step between whitening and coloring, which could be seen as \textit{style-construction}. We aim to transfer the style of a real cataract surgery to a simulation video, and to that end, the style of a single image is not representative enough of the whole surgery. Our approach performs a high-order decomposition of multiple-styles, and allows to linearly combine them by weighting their representations. On top of this, we introduce label-to-label style transfer by manually segmenting few images in the cataract challenge and using them to transfer anatomy style correctly. This is done by exploiting the fact that simulation segmentation masks can be extracted automatically, by tracing back the texture to which each rendered pixel belongs  \cite{zisimopoulos2017can}, and only few of the real cataract surgery have to be manually annotated. An overview of our approach can be found in \autoref{fig:overview}.

\vspace{-4mm}
\section{Proposed approach}

We formulate the multi-class multi-style transfer as a generalization to the recent work on
UST \cite{li2017universal}, which proposes a novel feed-forward 
formulation based on sequential auto-encoders to inject a given style into a content image by 
applying a WCT to the intermediate feature representation.
Our approach can further improve the alteration of the style blending aspects of the algorithm.

\subsection{Universal Style Transfer via WCT}

The UST approach proposes to address the style transfer problem as an image reconstruction 
process. Reconstruction is coupled with a deep-feature transformation to inject the style of interest into a given content image. To that end, a symmetric encoder-decoder architecture is built based on VGG-19 \citep{simonyan2014very}. Five different encoders are extracted from the pretrained VGG in ImageNet \cite{deng2009imagenet}, extracting information from the network at different resolutions, 
concretely after \textbf{relu\_x\_1} (for $x \in \{1, 2, 3, 4, 5\}$). Similarly, five 
decoders, each symmetric to the corresponding encoder, are trained to approximately reconstruct a 
given input image. The decoders are trained using the pixel reconstruction and feature reconstruction losses \cite{johnson2016perceptual,dosovitskiy2016generating}:
\begin{equation}
L = \norm{I_{in} - I_{out}}^2_2 + \lambda \norm{\Phi_{in} - \Phi_{out}}
\end{equation}
where $I_{in}$ is the input image, $I_{out}$ is the reconstructed image and $\Phi_{in}$ (as 
an abbreviation of $\Phi(I_{in})$) refers to the features generated by the respective VGG
encoder for a given input. After training the decoders to reconstruct a given image from the 
VGG feature representation (\textit{i.e.} find the reconstruction $\Phi(I_{in}) \rightarrow I_{in}$), 
the decoders are fixed and training is no longer needed. The style is transfered from one 
image to another by applying a transformation (e.g. WCT as described in the
next section) to the intermediate feature representation $\Phi(I_{in})$ and letting the  decoder reconstruct the modified features.

\subsubsection{Whitening and Coloring Transform}

Given a pair of intermediate vectorized feature representations $\Phi_c\in\mathcal{R}^{C\times H_sW_s}$ and $\Phi_s\in\mathcal{R}^{C\times H_sW_s}$, corresponding to a content $I_c$ and style $I_s$ images respectively, the aim of WCT is to transform $\Phi_c$ to approximate the covariance matrix of $\Phi_s$. To achieve this, the first step is to whiten representation of $\Phi_c$:
\begin{equation}\label{eq:1}
\Phi_{w}  = E_cD_c^{-\frac{1}{2}}E_c^T \,\, \Phi_c \,\,,
\end{equation}
where $D_c$ is a diagonal matrix with the eigenvalues and $E_c$ the orthogonal matrix of eigenvectors of the covariance $\Sigma_c = \Phi_c\Phi_c^T \in \mathcal{R}^{C\times C}$ satisfying $\Sigma_c = E_cD_cE_c^T$. After whitening, the features of $\Phi_c$ are decorrelated, which allows the coloring transform to inject the style into the feature representation $\Phi_c$:
\begin{equation}\label{eq:2}
\Phi_{cs} = E_sD_s^{\frac{1}{2}}E_s^T\,\,\Phi_{w}\,\,.
\end{equation}
Prior to whitening, the mean is subtracted from the features $\Phi_c$ and the mean of $\Phi_s$ is added to $\Phi_{cs}$ after recoloring. Note that this makes the coloring transform just the inverse of the whitening transform, by
transforming $\Phi_{wc}$ into the covariance space of the style image $\Sigma_s~=~\Phi_s\Phi_s^T = E_sD_sE_s^T$. The target image is then reconstructed by blending the original content representation $\Phi_c$ and the resultant stylized representation $\Phi_{cs}$ with a blending coefficient $\alpha$:
\begin{equation}\label{eq:3}
\Phi_{wct} = \alpha \,\, \Phi_{cs} + (1 - \alpha) \,\, \Phi_c.
\end{equation}
The corresponding decoder will then reconstruct the stylized image from $\Phi_{wct}$ after. For a given image, the stylization process is repeated five times (one per encoder-decoder pair).

\subsection{Generalized WCT (GWCT)}\label{sec:GWCT}

Although multiple styles could be interpolated using the original WCT formulation, by generating multiple intermediate stylized representations $\{\Phi^1_{wct},\dots,\Phi^n_{wct}\}$ and again, blending them with different coefficients, this would be equivalent to performing simple linear interpolation, which at the same time requires multiple stylized feature representations $\Phi_{wct}^i$ to be computed. Having a set of $N$ style images $\{I_s^1,\dots,I_s^n\}$, we first propagate them through the encoders to find their intermediate representations $\{\Phi^1_s,\dots,\Phi^n_s\}$ and from them, their respective feature-covariance matrices and stack them together $\bm\Sigma = \{\Sigma^1_s,\dots,\Sigma^n_s\} \in \mathcal{R}^{N\times C\times C}$. Then, the joint representation is built via tensor rank decomposition, also known as Canonical Polyadic decomposition (CP) \cite{kolda2009tensor}:
\begin{equation}\label{eq:4}
\bm\Sigma \,\, \approx \,\, \bm{P} = [[Z; Y; X]] = \sum_{r=0}^R \bm{z}_r \circ \bm{y}_r \circ \bm{x}_r,
\end{equation}
where $\circ$ stands for the Kronecker product and the stacked covariance matrices $\bm\Sigma$ can be approximately decomposed into auxiliary matrices $Z \in \mathcal{R}^{N\times R}$, $Y \in \mathcal{R}^{C\times R}$ and $X \in \mathcal{R}^{C\times R}$.

CP decomposition can be seen as a high-order low-rank approximation of the matrix $\bm\Sigma$ (analogous to 2D singular value decomposition (SVD), as used in the eigenvalue decomposition in equations \ref{eq:2} and \ref{eq:3}). The parameter $R$ controls the rank-approximation to $\bm\Sigma$, with the full matrix being reconstructed exactly when $R = min(N\times C, C\times C)$. Different values of $R$ will approximate $\bm\Sigma$ with different precision.

Once the low-rank decomposition is found (\textit{e.g.} via the PARAFAC algorithm \cite{kolda2009tensor}), any frontal slice $P_i$ of $\bm{P}$, which refer to approximations of $\Sigma^i_s$ can be reconstructed as:
\begin{equation}\label{eq:5}
\Sigma^i_s \,\, \approx \,\, P_i = YD^{(i)}X^T \quad \text{where} \quad D^{(i)} = \text{diag}(Z_i)
\end{equation}
Here $D^{(i)}$ is a diagonal matrix with elements from the column $i$ of $Z$. It can be seen that this representation encodes most of the covariance information in the matrices $Y$ and $X$, and by keeping them constant and creating diagonal matrices $D^{(i)}$ from columns $i$ of $Z$, with $i \in \{1,\dots,n\}$, original covariance matrices $\Sigma^i_s$ can be recovered.

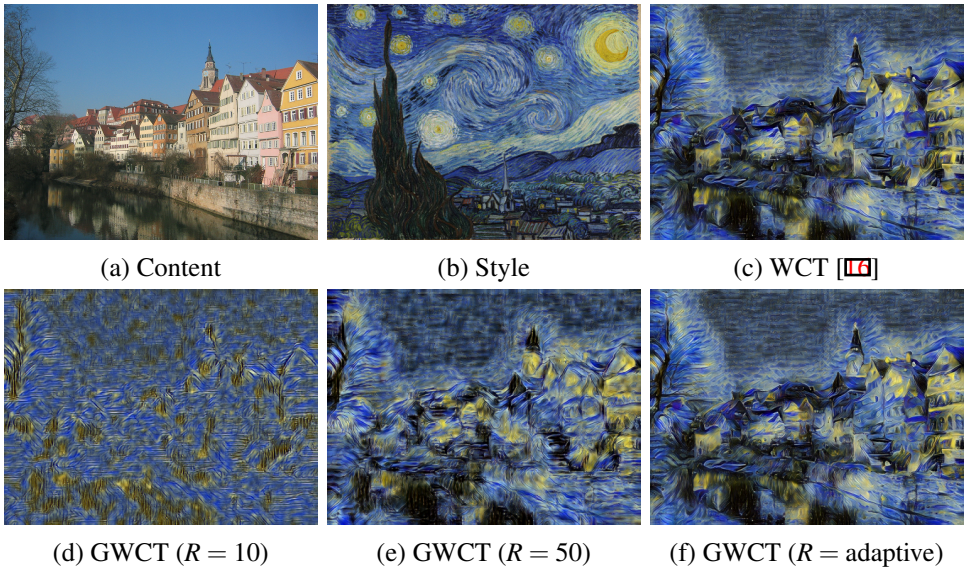
\begin{figure}[t]
\input{figure_rec.tex}
\vspace{-7mm}
\caption{GWCT as a generalization of WCT. Having the style model of $N$ images (as shown in section \ref{sec:GWCT}) with different low-rank approximations, one of the style covariances can be approximately reconstructed (with precision proportional to the rank $R$). (a) content, (b) style and (c) style transfer with WCT \cite{li2017universal}. (d), (e)  and (f) are low-rank approximations.}
\label{fig:reconstruction}
\vspace{-3mm}
\end{figure}

In order to transfer a style to a content image, during inference, the content image is propagated through the encoders to generate $\Phi_w$ (as in \autoref{eq:1}). Then, a covariance matrix $\Sigma^s_s$ is reconstructed from \autoref{eq:5}. The reconstructed covariance $\Sigma_w$ can then be used to transfer the style, after eigen-value decomposition, following \autoref{eq:2} and \autoref{eq:3} and propagating it through the decoder to obtain the stylized result.

\subsection{Multi-style transfer via GWCT}

From \autoref{eq:5} it can be seen that columns of $Z$ encode all the scaling and parameters needed to reconstruct covariance matrices. We can then apply style blending directly in the embedding space of $Z$ and reconstruct a multi-style covariance matrix.

Consider a weight vector $W \in \mathcal{R}^{N}$ where $W$ is $\ell_1$ normalized, then a blended covariance matrix can be reconstructed as:
\begin{equation}\label{eq:6}
\Sigma_w = YD^{(w)}X^T \quad \text{where} \quad D^{(w)} = \text{diag}(ZW).
\end{equation}
Here $D^{(w)}$ is a diagonal matrix where the elements of the diagonal are the weighted product of the columns in $Z$. When $W$ is a uniform vector, all the styles are averaged and, contrary, when $W$ is one-hot encoded, a single original covariance matrix is reconstructed, and thus, the original formulation of WCT is recovered. For any other $\ell_1$-normed and real valued $W$, the styles are interpolated to create a new covariance matrix capturing all their features.

As in the previous section, the reconstructed styled covariance from \autoref{eq:6} can be used for style transfer to the content features, and propagate it through the decoders to generate the final stylized result.

\subsection{Label-to-label style transfer via GWCT}

\begin{figure}[t]
\input{figure_multilabel.tex}
\vspace{-7mm}
\caption{Image-to-Image vs Label-to-Label (L2L) image stylization. (a) our simulation image, (b) the style Cataract image and (c) shows a WCT image-to-image style transfer. (d) and (e) represent content and style masks used to perform L2L stylization of (a) in (f).}
\label{fig:multilabel}
\vspace{-4mm}
\end{figure}
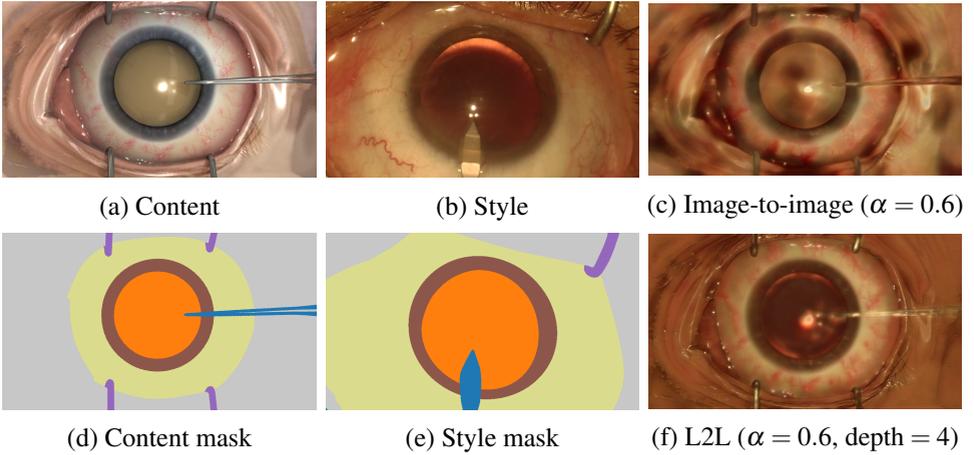

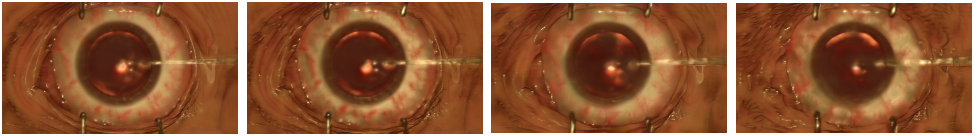
\begin{figure}[h]
\input{figure_multilabel2.tex}
\vspace{-7mm}
\caption{Effect of different hyper-parameters in Label-to-label stylizitation, using the same content/style images as in \autoref{fig:multilabel}}
\label{fig:multilabel2}
\vspace{-3mm}
\end{figure}

In our particular application, style transfer from real surgery to simulated surgery, additional information is needed to properly transfer the style. In order to be able to recreate realistic simulations the style, both color and texture, have to be transferred from the source image regions to the corresponding target image regions. Therefore, we define label-to-label style transfer as multi-label style transfer within a single image. Consider the trivial case were a content image and a style image are given, along with their corresponding segmentation maps $M$ where $m_i \in \{1,\dots,L\}$ indicates the class of the pixel $i$. Label-to-label style transfer could be written as a generalization of WCT, where the content and the style images are processed through the network and after encoding them, individual covariances $\{\Sigma^1,\dots, \Sigma^L\}$ are built by masking all the pixels that belong to each class. In practice, however, we aim to transfer the style to a video sequence and not all the images can contain all the same class labels than a single style image. This is, in our example of Cataract Surgery, multiple tools are used through the surgery and due to camera and tool movements, it is unlikely that a single frame will contain enough information to reconstruct all the styles appropriately. Our generalized WCT, however, can handle this situation inherently. As the style model can be built from multiple images, if some label is missing in any image, other images in the style set will compensate for it. The weight vector $W$ that blends multiple styles into one is then separated into per-class weight vectors $W^(i)$ with $i \in \{1,\dots,L\}$. We then can encode $W$ in a way that balances class information per image $W^i = C_j^i / \norm{C_j}_1$, where $N$ is the number of images used to create the style model, superscript indicate class label and subscript indicate the image index. $C_j^i$ then defines the number of pixels (count) of class $i$ in the image $j$. This weighting ensures that images with larger regions for a given class have more importance when transferring the style of that particular class.

\vspace{-5mm}
\section{Experimental Results}

\subsection{GWCT as a low-rank WCT approximation}

To validate the generalization of our approach over WCT, we conduct an experiment to prove that the result of WCT stylization can be approximated by our method. We first select four different styles and use them to stylize an image using WCT. We then build three different low-rank style models with them, with ranks $R = 10$, $R = 50$ and $R = adaptive$ respectively, as shown in section \ref{sec:GWCT}. $R = adaptive$ refers to the style decomposed with rank equal to the output channels of each encoder; this is, Encoder 1 outputs 64 channels and thus, uses rank $R = 64$ to factorize the styles, similarly, Encoder 5 outputs 512 channels resulting in a rank $R = 512$ style decomposition. After style decomposition, a low-rank approximation of each of the original styles is built from \autoref{eq:4} and used to stylize the content image. This process is shown in \autoref{fig:reconstruction} where the stylized image from WCT can be approximated with precision proportional to the rank-factorization of the styles. When $R = adaptive$, as explained above, our style transfer results and WCT are visually indistinguishable, proving our generalized formulation. Furthermore, the original style covariance matrices can be reconstructed exactly when $R = min(NC, CC)$ \cite{kolda2009tensor}. Also, in all our experiments $N \ll C$, which makes $C$ a sensible balance between computational complexity and reconstruction error. In all our experiments, unless stated otherwise, we choose $R = adaptive$. Here we should note that, different to the WCT, our approach does not require to propagate the style images through the network during inference and the style transforms are injected at the feature level. Style decompositions can be precomputed offline, and the computational complexity of transfering N or 1 style is exactly the same, reducing a lot the computational burden of transfering style to a video.

\begin{figure}[t]
\includegraphics[width=\textwidth]{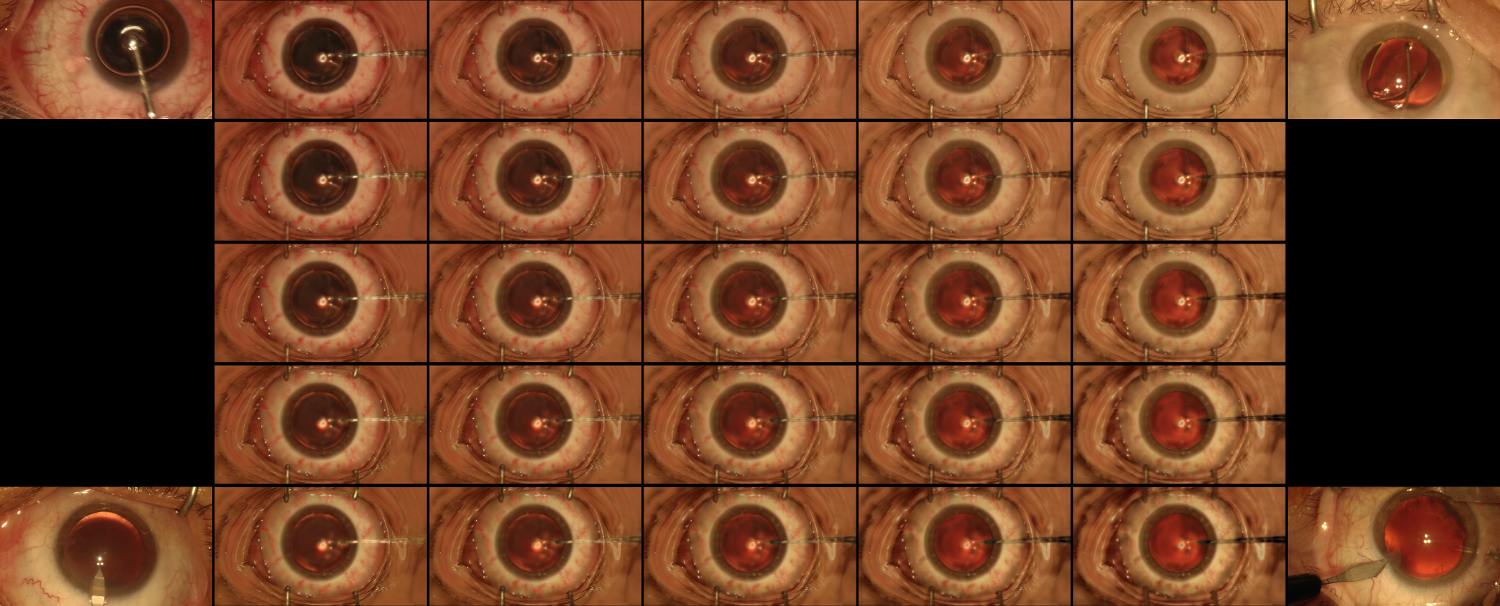}
\vspace{-7mm}
\caption{GWCT for multi-style interpolation. Four corners contain four different real cataract surgery images. The center $5\times 5$ grid of images correspond to the images of the simulated eye in \autoref{fig:multilabel} after interpolated style transfer.}
\label{fig:4way_collage}
\vspace{-4mm}
\end{figure}

\subsection{Label-to-label style transfer}

We show the differences between image-to-image style transfer and our GWCT with multi-label style transfer in \autoref{fig:multilabel} and \autoref{fig:multilabel2}. For these experiments different values of alpha $\alpha\in\{0.6, 1\}$ were used and of the maximum-depth of style encoding $\text{depth} \in \{4, 5\}$ are compared. $\text{Depth}$ refers to the encoder depth in which the style is going to start transferring (as per \autoref{fig:overview}). $\text{depth} = 5$, which means that the Encoder5/Decoder5 will be used to initially stylize the image and it will go up to Encoder1/Decoder1. However, if $\text{depth}$ is set to anything smaller $1 \leq \text{depth}\leq 5$, for example 4, then the initial level will be Encoder4/Decoder4, and pass through all of them until Encoder1/Decoder1. This means that different values of $\text{depth}$ will stylize the content image with different levels of abstraction. The higher the value, the higher the abstraction.

It can be seen in \autoref{fig:multilabel} and \autoref{fig:multilabel2} that, as previously mentioned, image-to-image style transfer is not good enough to create more realistic-looking eyes. By transferring the style from label-to-label, the style is transferred with much better visual results. Additionally the difference between $\text{depth} = 5$ and $\text{depth} = 4$ shows that sharper details can be reconstructed with a lower abstraction level. Images seem over-stylized with $\text{depth} = 5$. Having to limit the depth of the style encoding to the fourth level could be seen as an indicator that the style (or high-level texture information) is not entirely relevant, or that there is no enough information to transfer the style correctly.

\textbf{Label-to-label multi-style interpolation}: We show the capabilities of our GWCT approach to transfer multiple styles to a given simulation images using different style blending $W$ parameters in \autoref{fig:4way_collage}. Four real cataract surgery images are positioned in the figure corners. The central $5\times 5$ grid contains the four different styles interpolated with different weights $W$. This is, the four corners have weights $W = \text{onehot}(i)$, so that each one is stylized with the $i$-th image, for $i \in \{1,2,3,4\}$. The central image in the grid is stylized by averaging all four styles $W = [0.25, 0.25, 0.25, 0.25]$ and every other cell has a $W$ interpolated between all the four eyes proportional to their distance to them. The computational complexity of GWCT to transfer one or the four styles is exactly the same, as the only component that differs from one to the other is $D^{(w)}$ computation.

For this experiment the content image was selected to be a simulation image, as in the previous experiment, $\alpha = 0.6$ was selected for all the multi-style transfers, styles were decomposed with $R = adaptive$ and $\text{depth} = 4$ as it did experimentally provide more realistic transfers in this particular case. It can be seen that the simulated eyes in the corners accurately recreate the different features of the real eye, particularly the iris, eyeball and the glare in the iris. It is interesting to see how the different blending coefficients affect the multi-style transfers, as the style transition is very smooth from one corner to another, highlighting the robustness of our algorithm.

\subsection{Making simulations more realistic}\label{sec:simulations}

\begin{figure}[t]
\input{figure_end.tex}
\vspace{-7mm}
\caption{Style transfer from real Cataract surgery to video simulation.}
\label{fig:video}
\vspace{-4mm}
\end{figure}
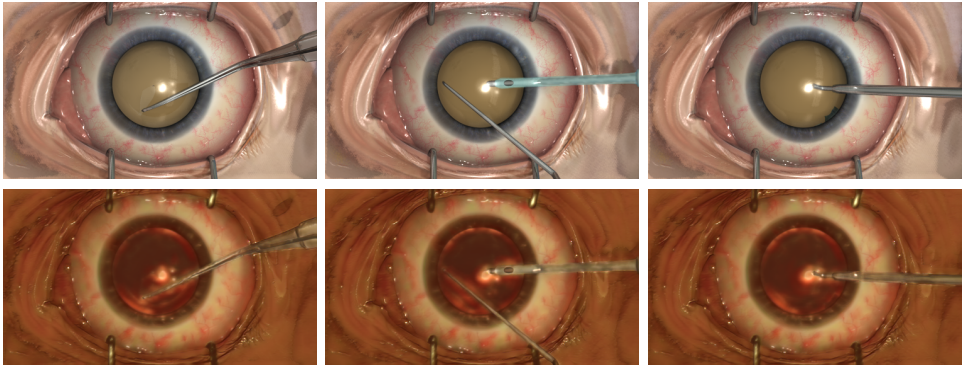

Finally, we prove our concept by transferring the style from a Cataract video to a real Video simulation. To that end we manually annotated (as in previous sections) the anatomy and the tools of 20 images from one of the Cataract Challenge. We have chosen only one of the videos to make sure that the style is consistent in the source simulation. All the Cataract surgery images are used to build a style model that then is transferred to the simulation video. Segmentation masks are omitted (due to lack of space). In order to achieve a more realistic result, we made $\alpha$ a vector to be able to choose different $\alpha$ values for each of the segmentation labels, using $\alpha = 0.8$ for iris, cornea and skin, $\alpha = 0.5$ for the eye ball and $\alpha = 0.3$ for the tools. Results are visible in \autoref{fig:video}. Full stylized simulation video is available in the supplementary material.

\section{Conclusions}

A novel method is proposed in this work to make surgical simulations more realistic, based on style transfer. Our approach builds on top of WCT and adds tensor decomposition and label-to-label style transfer to improve the style mapping from a reference surgical video to each of the various anatomical parts of our simulation. We show that style transfer is a powerful tool to improve the photo-realistic fidelity of simulations, and we pave the way towards using these results to generate large amounts of training data from these simulations, reducing the necessity of tedious and time-consuming manually annotated datasets. We believe our approach, and future work to come, could change how we create training datasets and it could speed up the data collection, particularly in fields where access to real-life surgical content is limited and difficult to capture.

\section{Acknowledgements}

We gratefully acknowledge the work of our Studio and Innovation team at Digital Surgery, particularly of Robert Joosten who generated the segmentation masks from the Cataract Simulation and our internal team of Rotoscopers, Nunzia Lombardo and Ellen Jaram who segmented the real Cataract images for us.

Danail Stoyanov receives funding from the EPSRC (EP/N013220/1, EP/N022750/1, EP/N027078/1, NS/A000027/1), Wellcome/EPSRC Centre for Interventional and Surgical Sciences (WEISS) (203145Z/16/Z) and EU-Horizon2020 (H2020-ICT-2015-688592).

\bibliography{paper.bib}
\end{document}

%% file: figure_rec.tex
\begin{tikzpicture}[every node/.style={
node distance=.33\textwidth, minimum width=.32\textwidth}]
\node[label=below:{(a) Content}] (a)
    {\includegraphics[width=.32\textwidth]{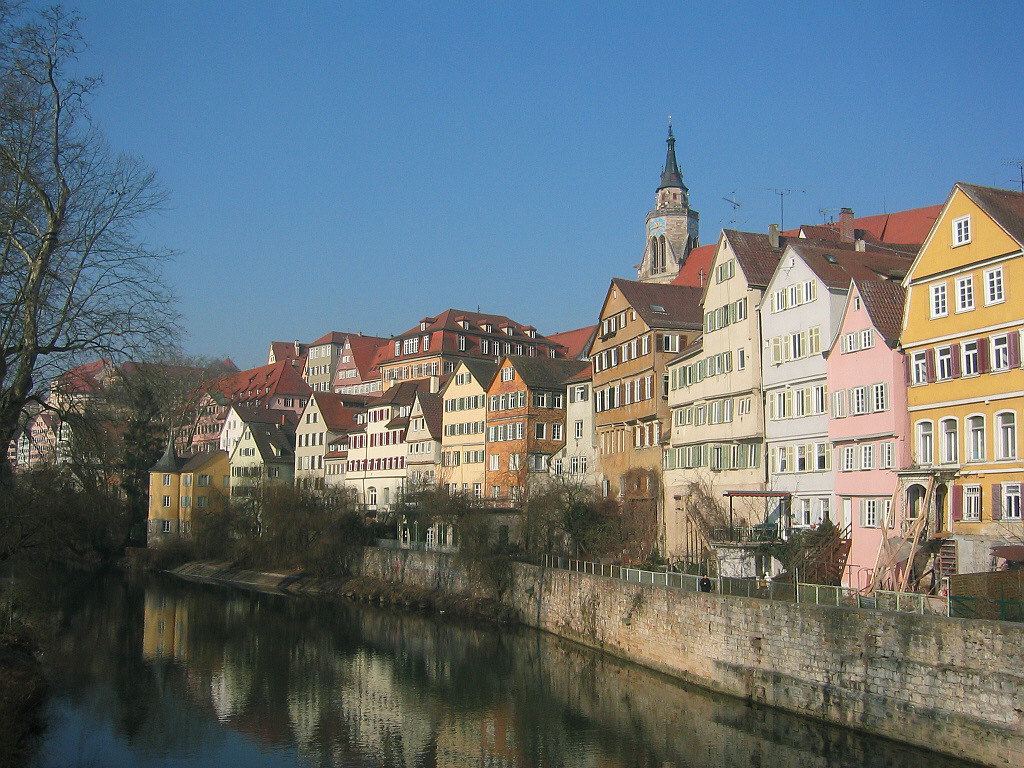}};
\node[label=below:{(b) Style}, right of=a] (b)
    {\includegraphics[width=.32\textwidth]{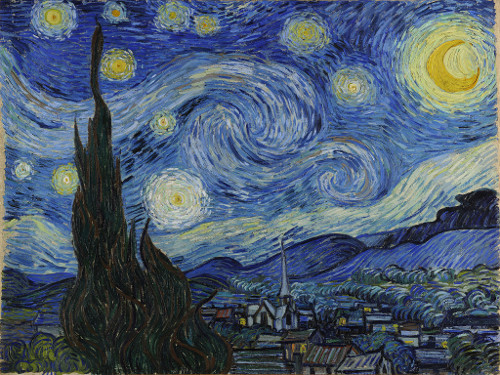}};
\node[label=below:{(c) WCT \cite{li2017universal}}, right of=b] (c)
    {\includegraphics[width=.32\textwidth]{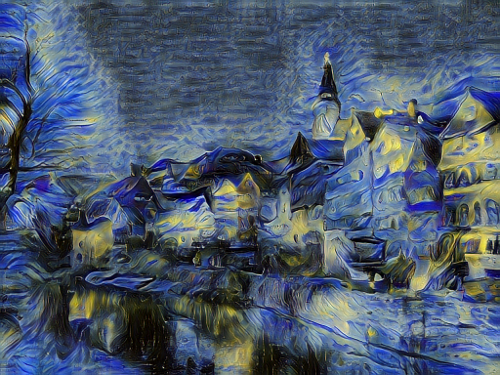}};

\node[label=below:{(d) GWCT ($R = 10$)}, below of=a, yshift=5mm] (d)
    {\includegraphics[width=.32\textwidth]{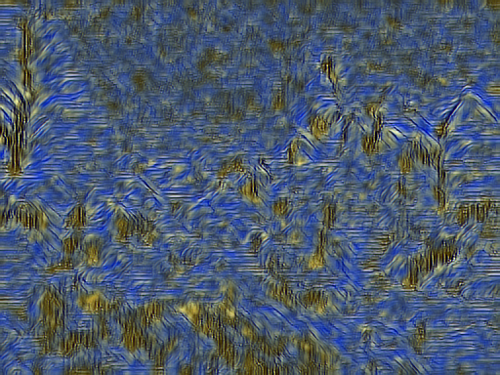}};
\node[label=below:{(e) GWCT ($R = 50$)}, right of=d] (e)
    {\includegraphics[width=.32\textwidth]{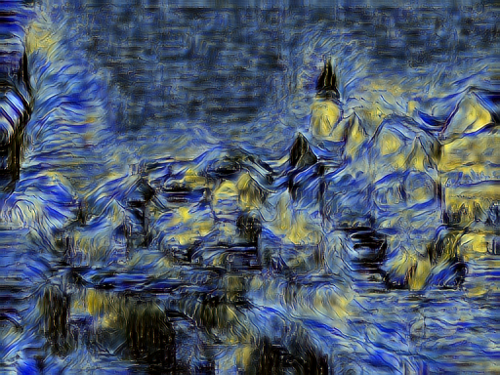}};
\node[label=below:{(f) GWCT ($R = \text{adaptive}$)}, right of=e] (f)
    {\includegraphics[width=.32\textwidth]{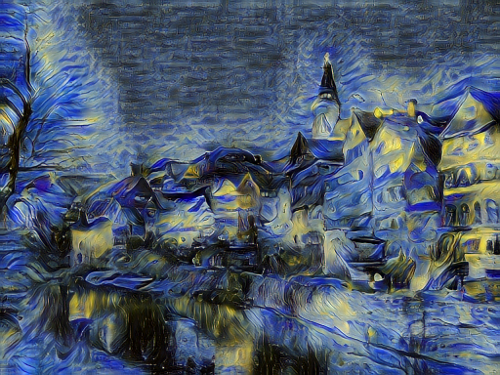}};
    
\end{tikzpicture}

%% file: figure_multilabel.tex
\begin{tikzpicture}[every node/.style={
node distance=.33\textwidth, minimum width=.32\textwidth}]
\node[label=below:{(a) Content}] (a)
    {\includegraphics[width=.32\textwidth]{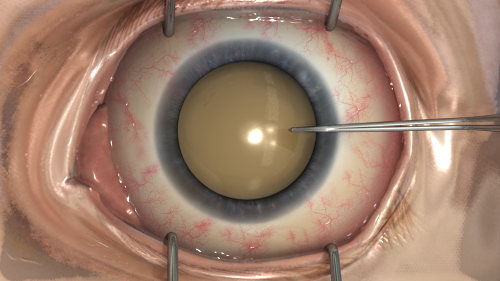}};
\node[label=below:{(b) Style}, right of=a] (b)
    {\includegraphics[width=.32\textwidth]{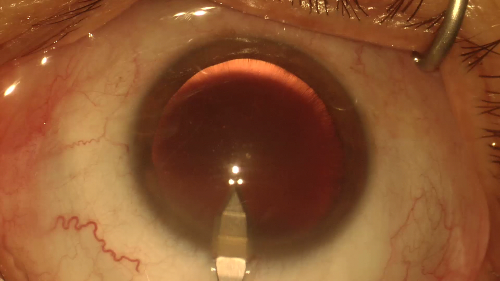}};
\node[label=below:{(c) Image-to-image ($\alpha = 0.6$)}, right of=b] (c)
    {\includegraphics[width=.32\textwidth]{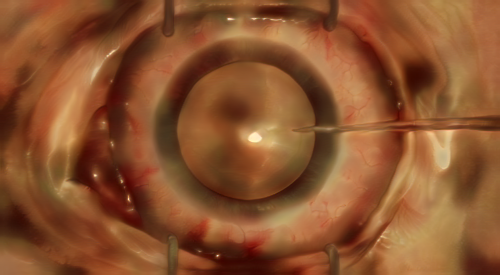}};

\node[label=below:{(d) Content mask}, below of=a, yshift=12mm] (d)
    {\includegraphics[width=.32\textwidth]{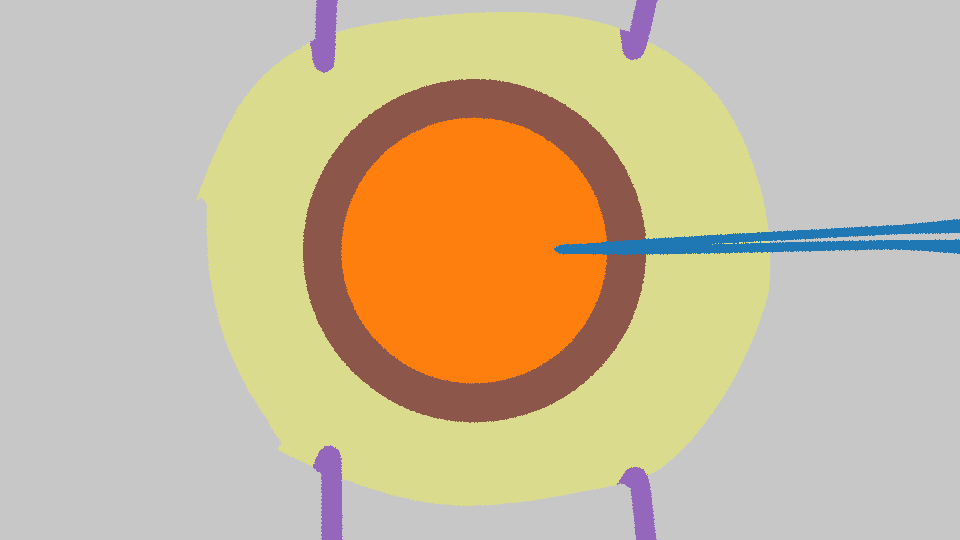}};
\node[label=below:{(e) Style mask}, right of=d] (e)
    {\includegraphics[width=.32\textwidth]{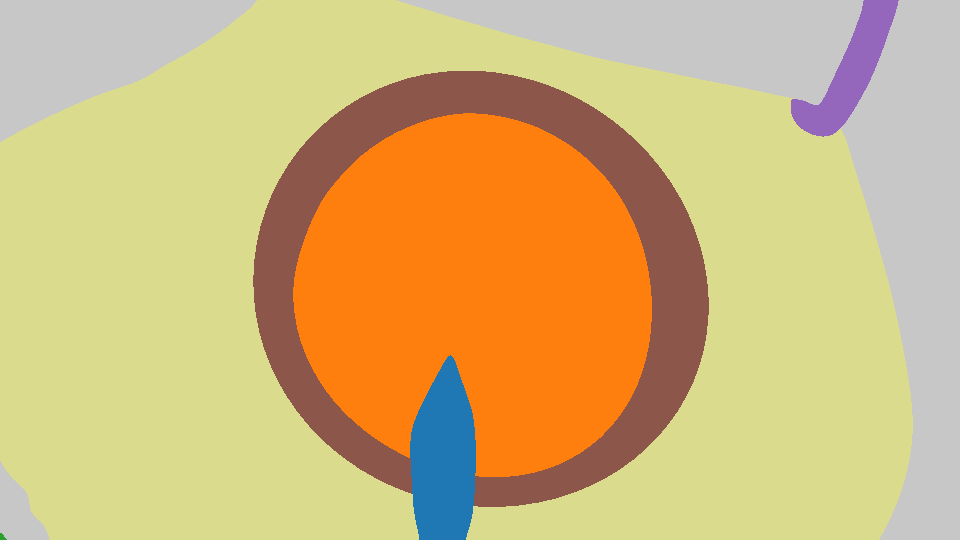}};
\node[label=below:{(f) L2L ($\alpha = 0.6$, $\text{depth} = 4$)}, right of=e] (f)
    {\includegraphics[width=.32\textwidth]{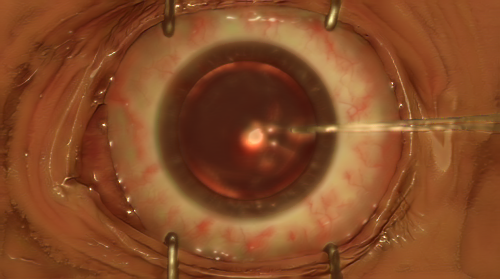}};

\end{tikzpicture}

%% file: figure_multilabel2.tex
\begin{tikzpicture}[every node/.style={
node distance=.25\textwidth, minimum width=.24\textwidth}]

\node[label=below:{(a) $\alpha = 0.6$, $\text{depth} = 4$}] (a)
    {\includegraphics[width=.24\textwidth]{stylized_ml_a0_6_d4_sd.png}};
\node[label=below:{(b) $\alpha = 1$, $\text{depth} = 4$}, right of=a] (b)
    {\includegraphics[width=.24\textwidth]{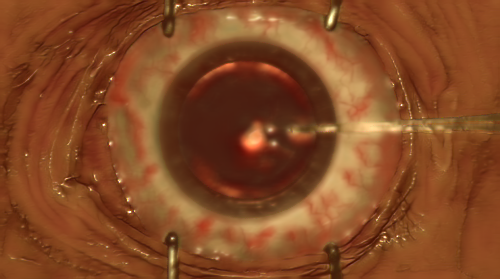}};
\node[label=below:{(c) $\alpha = 0.6$, $\text{depth} = 5$}, right of=b] (c)
    {\includegraphics[width=.24\textwidth]{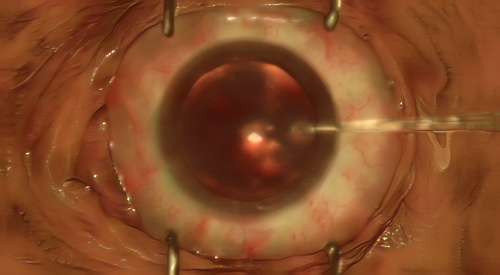}};
\node[label=below:{(d) $\alpha = 1$, $\text{depth} = 5$}, right of=c] (d)
    {\includegraphics[width=.24\textwidth]{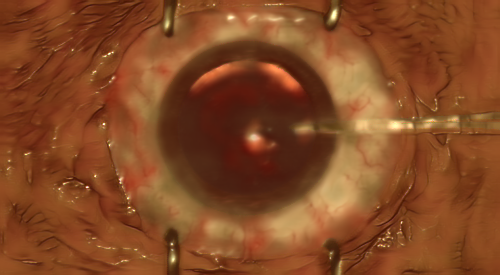}};
    
\end{tikzpicture}

%% file: figure_end.tex
\begin{tikzpicture}[every node/.style={
node distance=.33\textwidth, minimum width=.32\textwidth}]
\node (a)
    {\includegraphics[width=.32\textwidth]{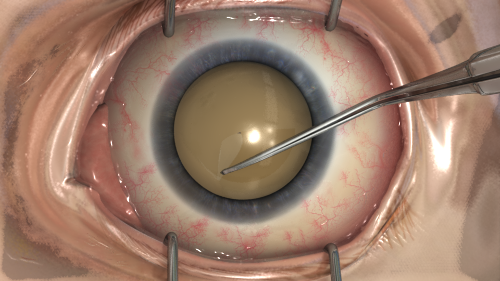}};
\node[right of=a] (b)
    {\includegraphics[width=.32\textwidth]{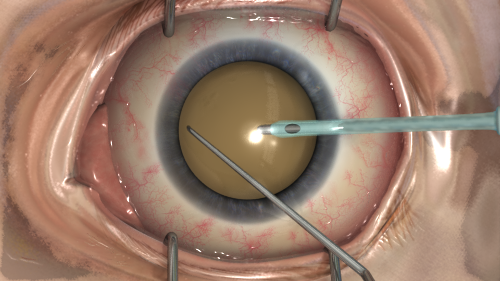}};
\node[right of=b] (c)
    {\includegraphics[width=.32\textwidth]{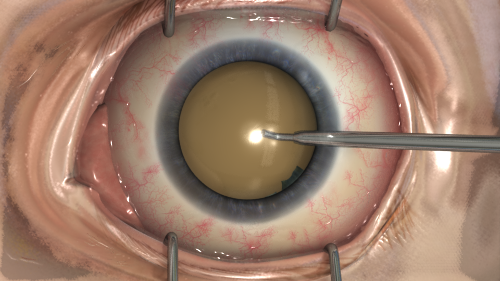}};
    
\node[below of=a, yshift=18mm] (d)
    {\includegraphics[width=.32\textwidth]{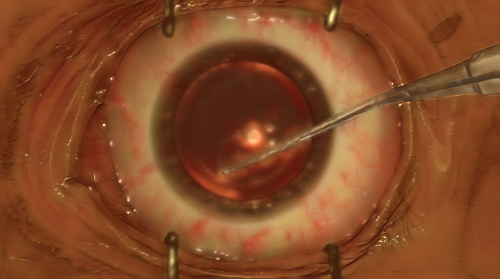}};
\node[right of=d] (e)
    {\includegraphics[width=.32\textwidth]{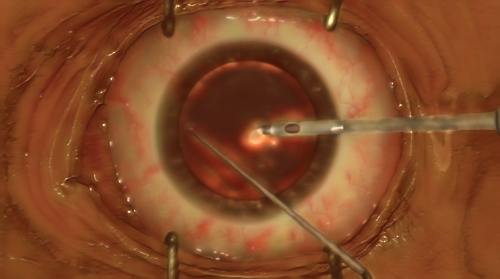}};
\node[right of=e] (f)
    {\includegraphics[width=.32\textwidth]{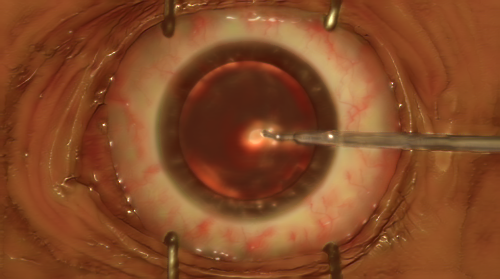}};

\end{tikzpicture}